\newif\ifconfpaper
\confpapertrue 

\newif\ifconfletter

\ifconfpaper
\documentclass[letterpaper, 10 pt, conference]{ieeeconf}
\IEEEoverridecommandlockouts
\fi

\ifconfletter
\documentclass[letterpaper, 10 pt, journal, twoside]{IEEEtran}  
\fi

\usepackage{graphicx}
\usepackage{amsmath}
\usepackage{amsfonts}
\usepackage{bm}
\usepackage{booktabs}

\newcommand{\secref}{Section~\ref}
\newcommand{\figref}{Figure~\ref}

\ifconfletter
\begin{document}
\fi
%
\ifconfletter
\title{The Title}
\else
\title{\LARGE \bf Task-sequencing Simulator: Integrated Machine Learning to Execution Simulation for Robot Manipulation}
\fi
%
%
%

\ifconfletter
\author{Kazuhiro Sasabuchi$^{1}$, Naoki Wake$^{1}$, and Katsushi Ikeuchi$^{1}$%
\thanks{Manuscript received: Month, dd, year; Revised Month, dd, year; Accepted Month, dd, year.}
\thanks{This paper was recommended for publication by Editor XXX upon evaluation of the Associate Editor and Reviewers' comments.
}
\thanks{$^{1}$All authors are with Applied Robotics Research, Microsoft, Redmond, WA, USA
        {\tt\footnotesize Kazuhiro.Sasabuchi@microsoft.com}}%
\thanks{Digital Object Identifier (DOI): see top of this page.}
}
\else
\author{Kazuhiro Sasabuchi$^{1}$, Daichi Saito$^{1}$, Atsushi Kanehira$^{1}$, Naoki Wake$^{1}$, Jun Takamatsu$^{1}$, Katsushi Ikeuchi$^{1}$
\thanks{$^{1}$All authors are with Microsoft, Redmond, WA, USA
        {\tt\small Kazuhiro.Sasabuchi@microsoft.com}}%
}
\fi
%
%

\ifconfletter
\markboth{IEEE Robotics and Automation Letters. Preprint Version. Accepted Month, year}
{Sasabuchi \MakeLowercase{\textit{et al.}}: The Title} 
\fi

%



\ifconfpaper
\begin{document}
\fi

\maketitle

\ifconfpaper
\thispagestyle{empty}
\pagestyle{empty}
\fi

\begin{abstract}
A task-sequencing simulator in robotics manipulation to integrate simulation-for-learning and simulation-for-execution is introduced.
Unlike existing machine-learning simulation where a non-decomposed simulation is used to simulate a training scenario,
the task-sequencing simulator 
runs a composed simulation using building blocks.
This way, the simulation-for-learning is structured similarly to a multi-step simulation-for-execution.
To 
compose both learning and execution scenarios,
a unified trainable-and-composable
description of blocks called a concept model is proposed and used.
Using the simulator design and concept models,
a reusable simulator for learning different tasks, a common-ground system
for learning-to-execution, simulation-to-real is achieved and shown.
\end{abstract}

\ifconfletter
\fi

%
\ifconfletter
\IEEEpeerreviewmaketitle
\fi

\section{Introduction}
\label{introduction}
%
%
%
%
\ifconfletter
\IEEEPARstart{T}{he} first sentence of the introduction.
\else
Simulators are important in robotics.
\fi
Compared to the real world,
simulators can run endlessly, safely, remove stochasticity, and provide ground truth data.
One direction for using simulators in robotics is to help check collisions and safe executions before real robot executions.
Another direction is using simulators as a tool for machine learning.
Simulators that suit either one of the above directions are available today,
however, it is desirable to have a simulator which fulfills both purposes to better integrate learning and execution,
especially in the multi-step manipulation domain.

In multi-step manipulation,
a series of tasks which occur sequentially must be simulated. 
An example of a sequential series of tasks is bringing an object, which is composed of tasks: grasp, pick, bring.
An execution simulator must be able to trigger these different tasks and connect them into a sequenced simulation.
When the simulator is used for checking robot executions,
the simulation is a matter of combining programmed or trained building blocks (e.g., run ``grasp" then ``pick" then ``bring").

In contrast,
machine learning simulators often ignore the task-sequence composition and
are structured to train a specific problem or benchmark (e.g., a non-decomposed ``pick-and-place" simulation)\cite{plappert2018multi}\cite{fan2018surreal}.
This structural difference causes a gap between simulation-for-execution and simulation-for-learning.
The learned results become specific to the trained scenario and contradicts with the simulation-for-execution where scenarios are non-fixed.

Instead, a machine learning simulator could be designed similar to an execution simulator.
The ``pick-and-place" scenario can be decomposed into a sequenced simulation of ``grasp then pick then bring then place then release."
The difference compared to the execution simulator is that some of these blocks are ``under-training" and are updated as data is collected.
Once the update has finished, the trained block can be combined and reused for a different scenario,
thus, the simulation-for-learning can directly transition to the simulation-for-execution.
In addition, such design enables learning new manipulation skills on top of programmed or prior-trained building blocks.
For example, a ``grasp" could be trained using the sequence ``grasp then pick," where ``pick" is a programmed task to provide a supervised signal
(i.e., teach that the ``grasp" was successful if the ``pick" was successful).

This article introduces a task-sequencing simulator structure 
which enables integrated learning-to-execution simulation.
At its core, the simulator uses a 
unified block design called the ``concept model,"
which is proposed within this article and defines the necessary descriptions
for training a task, collecting trained tasks, and running the tasks 
to compose a sequence.

   \begin{figure}[t]
      \centering
      \includegraphics[width=\columnwidth]{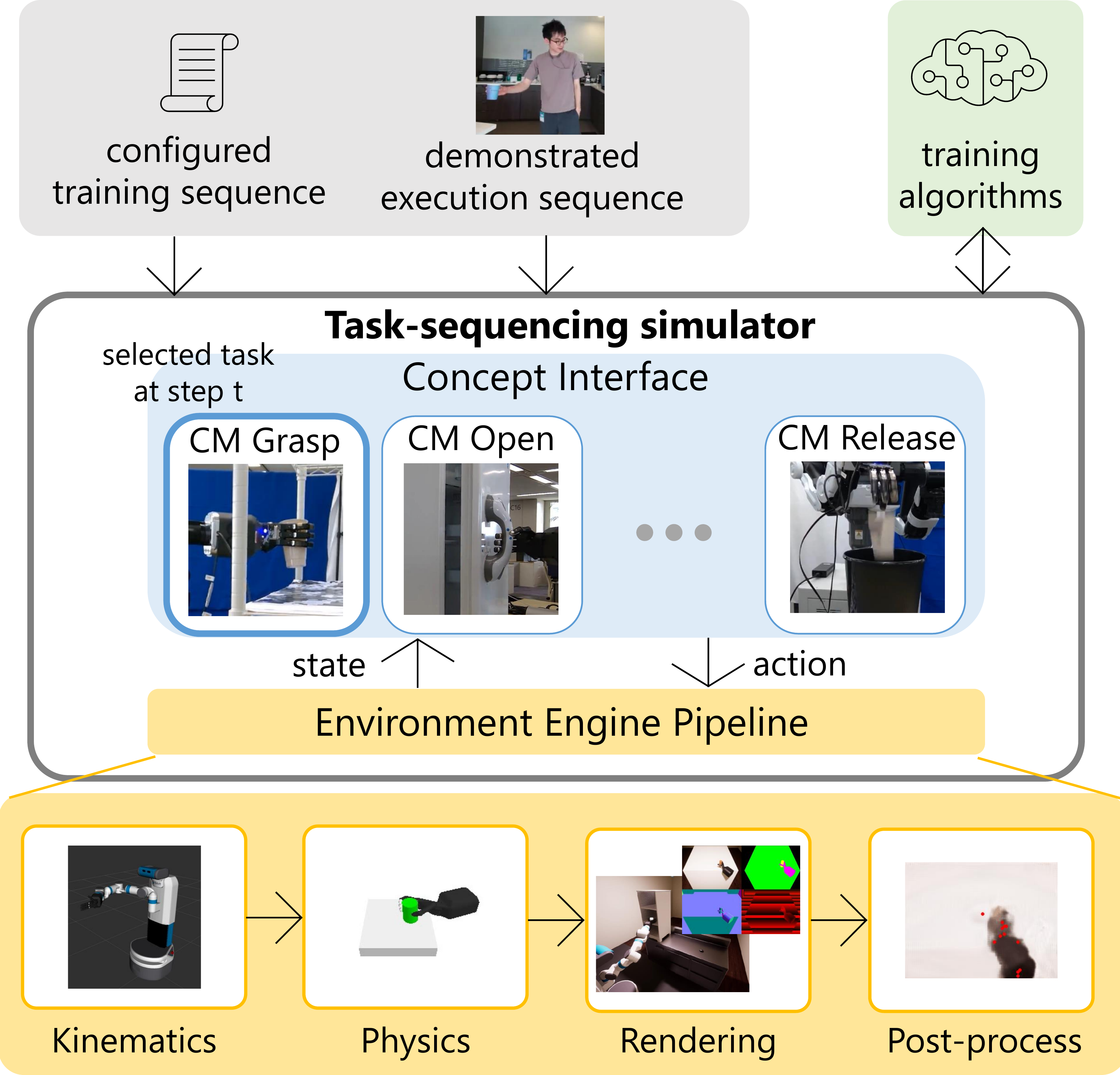}
	 \caption{The proposed task-sequencing simulator which enables scenario composition for both learning and execution in robotics manipulation.}
      \label{fig:main}
\vspace{-3mm}
   \end{figure}

The rest of the article is outlined as below:
\secref{works} provides a background on existing robotic simulators. 
\secref{sim} explains the overall simulator structure for achieving 
machine learning to robot execution simulation.
\secref{cm} explains the concept model core component of the simulator and \secref{impl} provides some detailed example implementations of the model.
\secref{exp} shows the capabilities of the simulator in machine-learning-to-robot-execution followed by conclusions in \secref{conclusions}.

\section{Background}
\label{works}


While there are many existing simulators for robotics,
existing simulators may not achieve the integrated learning-to-execution multi-step manipulation purpose for one of the following reasons:
1) the simulator targets a different domain other than manipulation,
2) the simulator can be used for manipulation but misses a capability in simulation-for-learning,
3) the simulator can be used for manipulation but misses a capability in simulation-for-execution,
4) the simulator can be used for manipulation and both learning-and-execution purposes but not specifically for learning-to-execution purposes.

Popular robotics simulators\cite{collins2021review}\cite{zhao2020sim}
include Gazebo\cite{koenig2004design}, MuJoCo\cite{todorov2012mujoco}, CoppeliaSim\cite{rohmer2013v}, CARLA\cite{dosovitskiy2017carla}, AirSim\cite{shah2018airsim}, and Webots\cite{michel2004cyberbotics}.
Gazebo has its advantage in its capability to simulate executions using ROS integrated sensors and actuators but is not the best choice when it comes to data collection and machine learning due to its slow simulation performance and inconsistency in physics simulation.
Thus, Gazebo falls into the second category.
Engines like MuJoCo on the other hand, are suitable for stable physics simulation in machine learning but miss some robotics simulation capabilities such as inverse-kinematics and visual feedback (realistic rendering). 
The focus is on physics simulation rather than an integrated simulator
for robot executions, therefore falls into the third category.
CARLA and Airsim mainly target automobiles such as drones and cars therefore miss some important features such as kinematics required for manipulation and falls into the first category.

The CoppeliaSim is an integrated simulator with a kinematics and physics engine, and the PyRep toolkit\cite{james2019pyrep} can be used with the simulator for machine learning.
WeBots is also an integrated simulator and frameworks such as Deepbots\cite{kirtas2020deepbots} help the simulator to be used for machine learning.
The machine learning features of these simulators are external features that have been developed within the community.
While it is possible to use these simulators for both execution and learning purposes,
they have not been designed for integrated learning-to-execution but rather using for one-or-the-other purpose.
That is, these simulators are not designed to connect learning-and-execution,
rather, learning and execution are separate use cases where one uses a community provided wrapper for machine learning,
and the other uses the integrated features to simulate a robotic system execution.


Compared to the existing simulators,
the task-sequencing simulator 
was designed to connect simulation-for-learning and simulation-for-execution
The simulator uses a concept model which
enables composition of pre-trained, programmed, or trained tasks, which is a powerful feature for going from machine learning to real robot execution.
(e.g., such as
plugging-in to machine learning platforms but then connecting to execute on ROS). More importantly, tied-integration allows 
features such as training using pre-sequent and post-sequent task executions, but also
collecting reusable execution modules through training.



\section{Task-sequencing Simulator Overview}
\label{sim}

The task-sequencing simulator has two layers: the Concept Interface for ``action decision" and the Environment Engine Pipeline for ``state observing" (\figref{fig:main}).
However, unlike a typical learning simulator, where
a specific problem has a non-decomposable structure and
the action decision is a single policy being updated as data is collected for the problem,
the task-sequencing simulator
adds an abstraction to this action decision so that
the problem is composed of a sequence of tasks (i.e., switches between a collection of tasks, where each task runs an individual policy),
This way,
a learned task policy 
can become part of 
a collection of policies for execution
once the training has finished.
Further details of each layer are described below.

\subsection{concept interface}
\label{ci}

At each simulation time-step, a robot decides the next action depending on the current state of the world.
This decision is referred to as a policy.
When the relation between the state, action, and next state (system dynamics) is completely known,
this policy can be directly programmed.
When the system dynamics are unknown, either the learning of the policy or system dynamics is required through data collection.
Data collection is efficient if collected only for the unknown dynamics and if known dynamics are directly computed.
Therefore, it is preferable to break down a robot's execution to a series of tasks,
where each task executes its own policy optimal for the system dynamics the task is covering.
In addition, breaking down a robot's execution increases the reusability of each task policy and allows composing
different execution scenarios from the task building blocks.

The Concept Interface layer 
chooses and switches
between the tasks for a training or execution scenario assuming
(1) the series of tasks to simulate is known (ways for knowing are shown in the experiments),
and (2) each task policy can indicate when the task has been completed (i.e., has a learned or programmed completion signal).
During simulation-for-execution, the Concept Interface layer chooses the according task in a sequence
and switches to the next task once the current policy returns a task completion signal.
The simulation-for-learning can be conducted in a similar way, except, the policy's output of the completion signal is evaluated
(e.g., by the success of a subsequent task).
This interchangeable structure
enables an integrated learning-to-execution simulation. 
To learn the termination signal under an arbitrary training sequence,
the tasks share a unified design called a ``concept model," which is
further explained in the later sections.

\subsection{environment engine pipeline}
\label{envg}

At a time-step control scale, a robot behaves under a sense-plan-act with a {\it physical embodiment}.
Thus, simulation in robotics requires three important engines:
kinematics, physics (contact dynamics), and rendering.
The role of the kinematics engine is to do the simulation between the robot's action plan and the movement of the actual body.
The role of the physics engine is to do the simulation between the robot's body and the environment.
The role of the rendering engine is to do the simulation between the environment and the robot's sensing.
In more technical terms,
the kinematics engine solves the mapping between cartesian space and configuration space,
the physics engine solves the differential algebraic equation\cite{haug1989computer} using techniques such as velocity-impulse linear complementarity-based time-stepping methods\cite{stewart1996implicit}\cite{anitescu1997formulating}, 
the rendering engine solves the rendering equation\cite{kajiya1986rendering} about lighting paths for pixel color generation.

While rendering engines simulate the robot's sensing by generating images,
there is a gap between sensing and perceiving (extracting meaningful states from the generated images).
Rather than directly learning the sensing-to-planning,
sometimes it is more efficient to perceive-before-planning and extract visual features.
Moreover, in robotics it is important to combine both visual and force feedback;
the visual features help compress the feedback so that the vision and force
have an aligned state dimension.
Thus, the proposed simulator adds a fourth ``post-process engine" in conjunction with the rendering engine.

These different-role engines are triggered in an ordered pipeline to calculate the current state of the simulation world.
Often, simulators package specific engines to produce a single simulation world, but in general,
these engines could run separately and combine/orchestrate multiple simulation worlds to produce better simulation.
For example, ROS MoveIt could be used for accurate inverse kinematics simulation,
PyBullet for reproducible physics simulation, and the Unreal Engine for photo-realistic ray-traced rendering.
Combining different engines is possible as long as each engine is able to load the same models of the robot/objects
and is able to share the robot and object states among each engine (which can be done using TCP connections etc.).
In addition, instead of using simulation engines, it is also possible to connect ``real" engines
which replace simulated physics with the real robot's torque sensors and simulated rendering with real images from the robot's camera.


\section{Concept Models}
\label{cm}

\begin{figure}[t]
      \centering
      \includegraphics[width=\columnwidth]{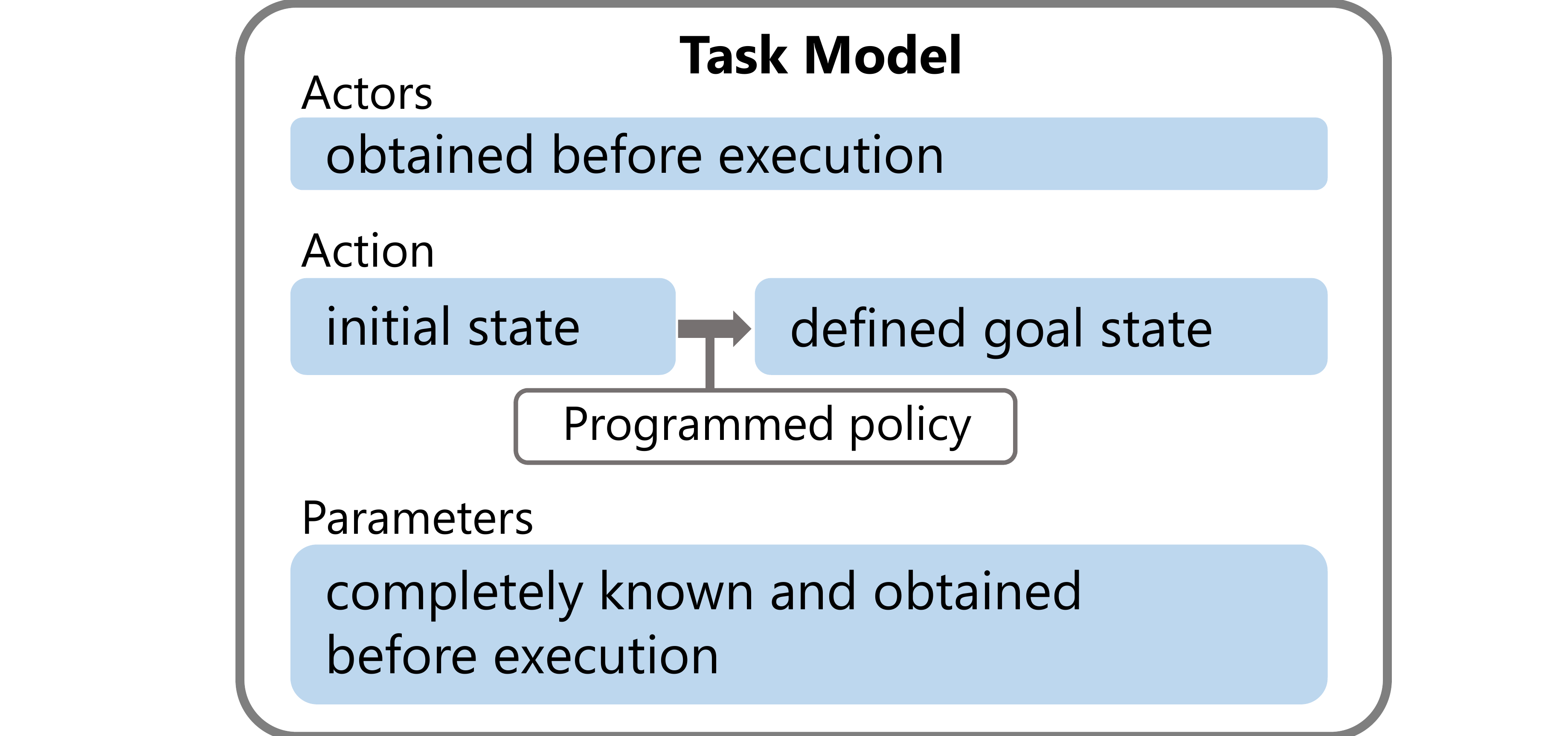}
	 \caption{An illustration of a task model.}
      \label{fig:tm}
   \end{figure}

A single task block operates under some system dynamics and achieves a goal state from an initial state.
Thus, the details of a task can be described using the actors of the system, an initial state, a goal (end) state, and the parameters of the system dynamics
(\figref{fig:tm}).
This kind of task description can be referred to as a ``task model"\cite{ikeuchi2021semantic}.
This description is enough for executing a task if the system dynamics are completely known.
The initial state is usually the end state of the previous task.
However, when the dynamics are not fully known, learning is required, and during learning, the initial state must be randomized.

\begin{figure}[t]
      \centering
      \includegraphics[width=\columnwidth]{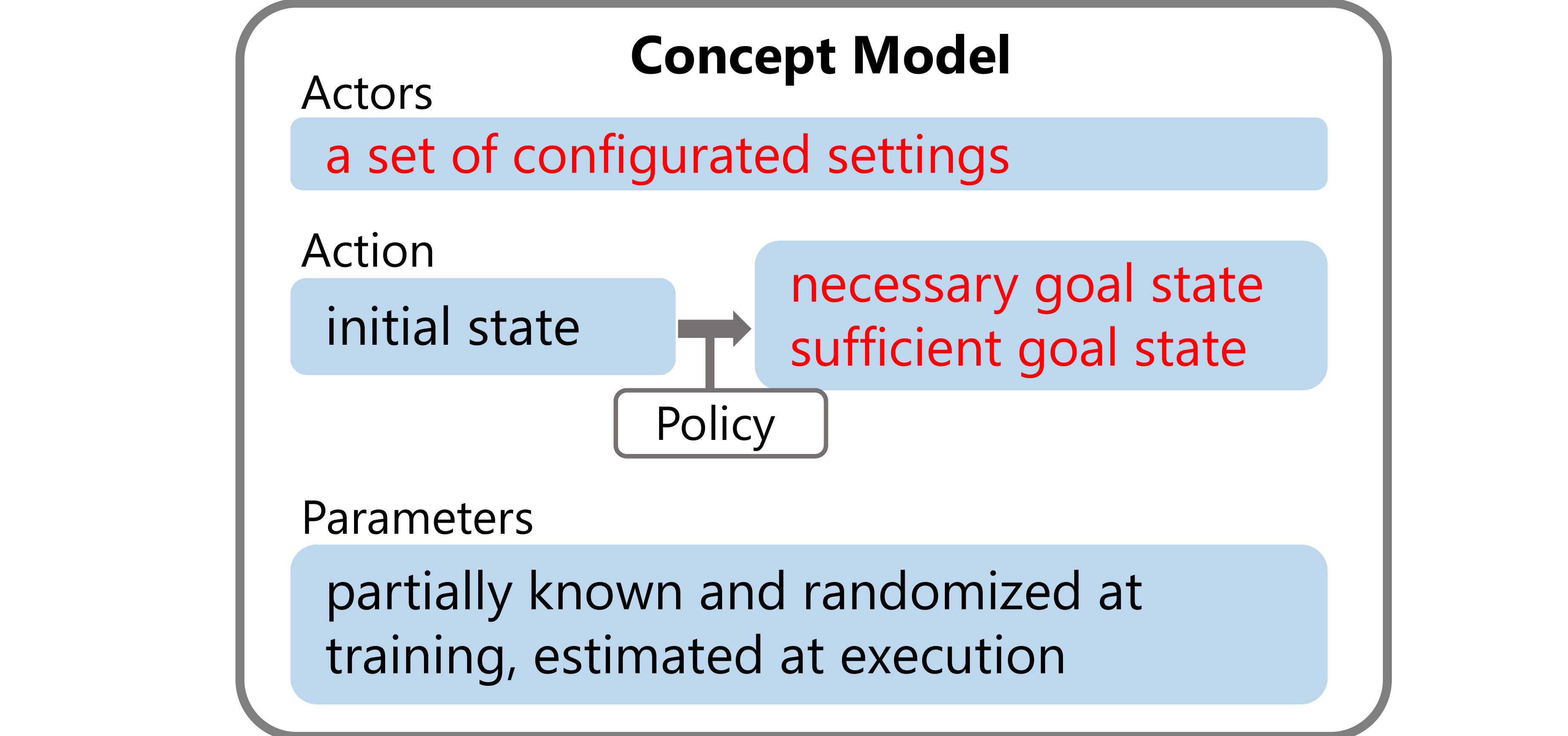}
	 \caption{An illustration of a concept model.}
      \label{fig:cm}
\vspace{-3mm}
   \end{figure}

Instead of fully describing the task, a task can be described using
actor configurations, an initial state, a necessary goal state that is described from observable system states,
a sufficient goal state that is described from non-observable system states, and the partially known parameters of the system dynamics
(\figref{fig:cm}).
This kind of task description will be referred to as a ``concept model" which compared to the task model may not be concrete enough for execution,
but describes the concepts of the task to learn the execution.
In the special case where the system dynamics are fully known and the task is programmable,
the descriptions of the concept model is identical to the task model.

By providing a structured description format,
the Concept Interface layer can access these blocks interchangeably
within a training sequence
as the structures are the same and only differ in the details.

\subsection{concept model usage in learning}

The concept model descriptions are used to learn the task completion signal as well as the actions to handle the unknown system dynamics.
The necessary goal state and sufficient goal state descriptions are used to evaluate whether the task completion signal and actions are appropriate.
The evaluation is done by minimizing the cost of the current states to the goal states.
Note that 
the cost to the necessary goal state is evaluated after every action decision,
whereas the cost to the sufficient goal state is only evaluated once a task completion signal is chosen during training.

The actor configurations describe the possible environments (the world including the robot and any manipulating target) for training the task.
If there are no pre-sequent tasks involved for training, then one of the actor configurations is used to define the initial state of the tasks.
Otherwise, the end state of the previous task is the initial state.
Unlike the initial state, the actor configuration is independent from the states of the previous task,
thus is configurable and can be used for randomizing the states for training.

\subsection{concept model usage in execution}

When the dynamics are fully known, the concept model acts the same as the task model.
The state of the task changes from the initial state using actions from a programmed policy based on the system dynamics and actors.
The initial state of the task during execution is the end state of the previous task
and the task ends once the goal state is achieved.

When the dynamics are not fully known,
the observable system states of the task changes using a learned policy.
Similar to the programmed case,
the initial state is the observable states at the end of the previous task.
However, since part of the goal state is non-observable,
the end of the task cannot be identified just with the model descriptions.
Instead, the learned task completion signal from the concept model descriptions is used to identify the end of the task. 



\section{Model Implementations}
\label{impl}

\begin{figure}[t]
      \centering
      \includegraphics[width=\columnwidth]{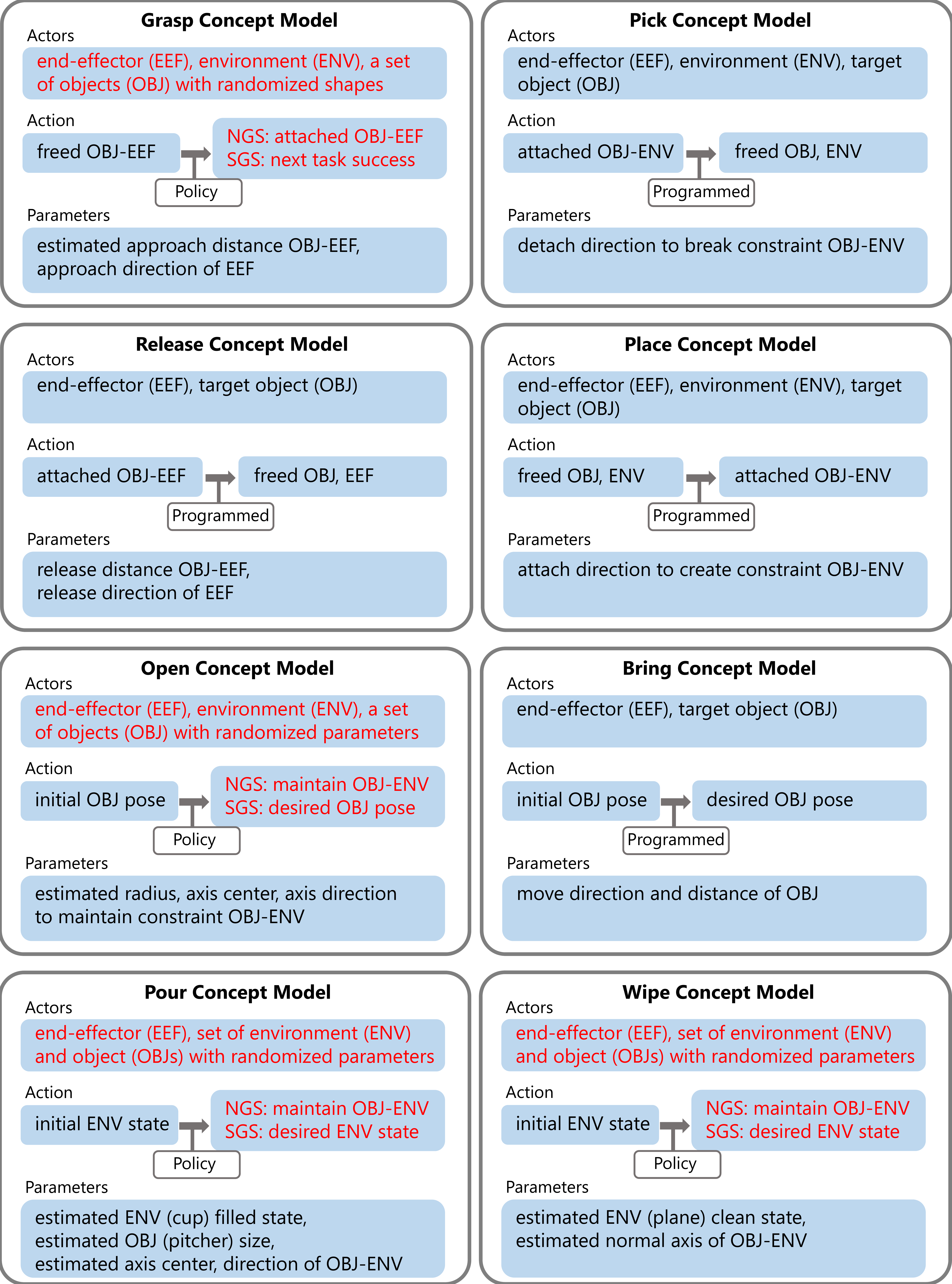}
	 \caption{Example concept models of eight different tasks in the screw-theory based classification.}
      \label{fig:models}
   \end{figure}

By following the concept model structure, a task is implemented in a way that can be trained but then collected as a building block for execution.
In this article,
the screw theory\cite{ohwovoriole1981extension} based separation of dynamics\cite{ikeuchi2021semantic} is used to separate a task from another task.
That is, once the relation between the manipulation target and the robot's end-effector is initialized,
a task breaks or maintains a contact state 
between the target and the environment.
By classifying the inequality equation patterns of contact points, this leads to seven pure translation tasks and seven pure rotation tasks.
\figref{fig:models} shows implementation examples of some of these tasks as concept models.
Below describes the details of some of the examples in the figure.
Note that as the task classification only depends on the relation between the end-effector, manipulation target, and the environment,
the movement of the arm (configuration space) can be ignored\cite{sasabuchi2020task}
and the task only focuses on the movement of the end-effector (cartesian space).

\subsection{grasping}

The grasp task initiates the relation between the manipulation-target and the robot's end-effector.
The actors are a target object, an environment (e.g., table), and the end-effector.
The initial state is where the target object is attached to the environment but not attached to the end-effector (including shape of the finger joints).
The goal state is where the target object is also attached to the end-effector in a way such that enough force is exerted for performing a subsequent task.
The parameters are the distance between the target and end-effector as well as the approaching direction of the end-effector.

When details of the target object are completely known, a task model can be defined and programmed from the above details.
However, in the real world, there is uncertainty in the shape of the object and distance to the target,
distinguishing whether enough force is exerted is intractable due to the inaccuracy in the real contact sensors or lack of sensors to detect slipping,
a grasp-failure due to finger-object collision during approach may occur if the policy is not carefully designed under the uncertainties of the object properties.

Instead, a set of actor configurations is defined as object shapes from a randomized range,
a necessary goal state is defined as the end-effector to be in contact with the target object on an appropriate surface
(which can be obtained on the real robot with the finger configurations and
finger-torque sensors with a threshold to determine a binary contacted-or-not-contacted state),
a sufficient goal state is defined as a successful performance of a subsequent task,
the estimated distance is used for the parameters.
The defined goal states are used to formulate the reward (cost-to-go) for learning the policy.

Using this concept model, the approaching strategy (adjusted movement around the approaching direction of the end-effector)
and the enough amount of ``closing" of the fingers to perform a subsequent task is learned.
The learned policy chooses the sufficient amount based on the object shape which can partially be inferred
by the shape of the end-effector finger joints once touching the object. 
The policy returns a termination signal once reached the enough amount of closing.

\subsection{door-opening}

The door-opening task is a one degrees-of-freedom pure rotation task.
The actors are a target object (the door), an environment (the hinge), and the end-effector (attached to the door handle).
The initial state is where the end-effector and target object are at an attached state.
The goal state is where the target object has moved to some desired orientation.
The parameters are the rotation radius, the rotation axis center, and the rotation axis direction defined by the target and environment.

When details of the target and environment are completely known, a task model can be defined and programmed from the above details.
However, in the real world, there is uncertainty in the environment parameters.
Instead, a set of actor configurations randomizing the radius,
a necessary goal state that moves the object along the environment constraint at each time-step
(which can be obtained on the real robot by using a force sensor on the wrist and checking against a maximum-stress threshold),
a sufficient goal state that ensures the door has reached the desired orientation,
and estimation of the parameters are used to describe the concept model.

By using an end-effector with only force-sensor feedback on the wrist,
this model enables learning a policy which updates parameter estimations at each time-step,
and then generates a hand motion based on the updated parameters.
The policy returns a termination signal once inferred that the desired orientation has been reached.

\subsection{bringing}

The bringing task is a six degrees-of-freedom translation and rotation task.
The actors are the target object and the end-effector.
The initial state is where the end-effector and target object are at an attached state.
The goal state is where the target object has moved to some relative positioning.
The parameters are the moving direction and distance.

Since there are no uncertainties in the target or environment,
the parameters can be manually specified and the goal can be directly specified from the parameters,
Thus, the concept model is identical to the task model and can be programmed.


   \begin{figure*}[t]
      \centering
      \includegraphics[width=2\columnwidth]{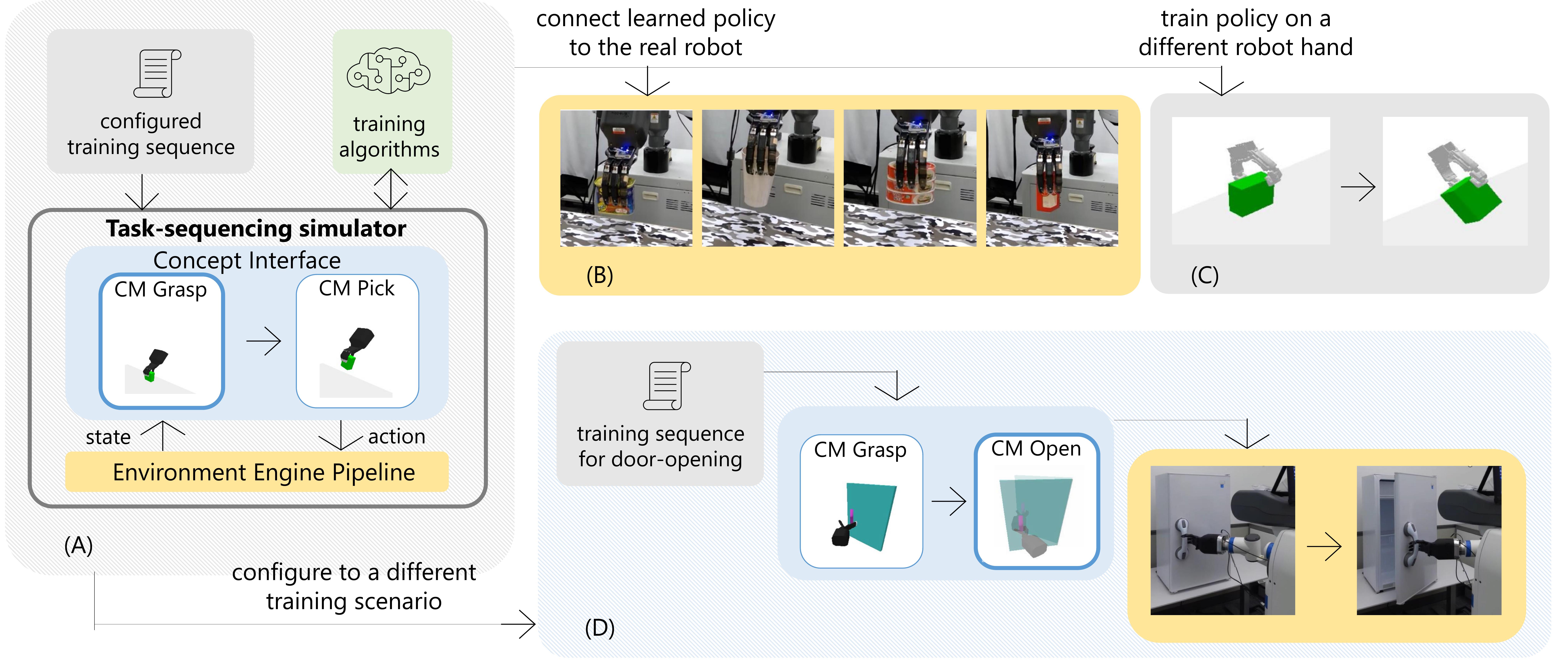}
	 \caption{Results of the task-sequencing simulator when used for learning.}
      \label{fig:ml}
   \end{figure*}

   \begin{figure*}[t]
      \centering
      \includegraphics[width=2\columnwidth]{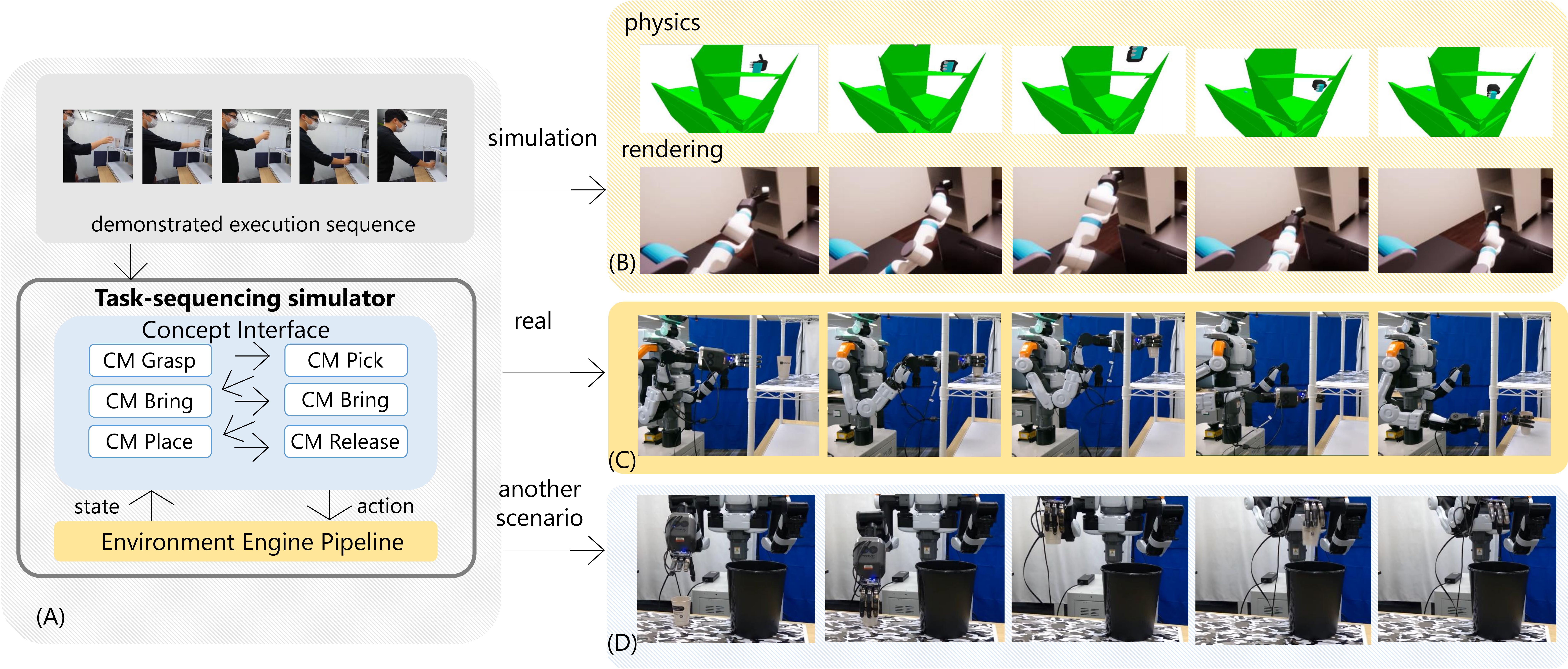}
	 \caption{Results of the task-sequencing simulator when used for execution.}
      \label{fig:re}
   \end{figure*}

\section{Experiments}
\label{exp}


Experiments were conducted
using the concept model implementations shown in the previous section,
and the developed task-sequencing simulator.

For the learning experiments,
the series of tasks to simulate were predefined. 
These experiments were performed to show the
effectiveness of the simulator and its reusability for training different tasks. 

By running the pre-sequent tasks of the task-to-train at the start of an episode,
and by running the subsequent tasks at reward return,
the simulator is compatible with common reinforcement learning platforms,
and utilizes off-the-shelf learning algorithms.
For the experiments,
the simulator was connected to the Bonsai platform and used the PPO algorithm.

For the execution experiments,
the series of tasks were obtained from human demonstrations and the
actions were generated using the learned task policies.
These experiments were performed to show the
effectiveness of the simulator for 
execution simulations (execution of different composed scenarios).
Note that the exact same simulator and task blocks used for training was used for this experiment,
showing the simulator's capability to transition from simulation-for-learning to simulation-for-execution.

For simulation,
states were obtained by plugging to the environment engine pipeline the PyBullet engine for the physics engine role,
and the Unreal Engine for rendering.
For the real robot, states were obtained plugging ROS (connected via roslibpy) to the environment engine pipeline.
For the arm kinematics, the ROS MoveIt package was used.

\subsection{training}

\figref{fig:ml}-(A) shows the task-sequencing simulator configurations for running the grasp training.
A configured training sequence ``grasp then pick" is passed to the simulator.
The ``grasp" task is the task-to-train, and a programmed ``pick" is used as the subsequent task for evaluating the sufficient goal state.

\figref{fig:ml}-(B) shows the trained grasp results performed on a real robot.
The concept model was designed with the object shape parameters as unknown.
Regardless of such uncertainty, the learned policy successfully grasps the different shaped objects including but not limited to
a box, a cylindrical cup, an oval rice pack, and a diamond-shaped candy box.

\figref{fig:ml}-(C) shows a learned grasp for a different robot hand,
which was trained using the same simulator and concept models but with a different actor configuration (end-effector) setting.
The results show the reusability of the simulator for training different robots with different mechanics
(a hand with multiple fingers and a gripper with limited degrees of freedom).

\figref{fig:ml}-(D) shows that by changing the configured training sequence to ``grasp then open,"
the door-opening task is trained using the same simulator.
The ``open" task is the task-to-train, and the trained ``grasp" is reused as the pre-sequent task for initiating the relation between the end-effector and the target door.
Regardless of uncertainty in the rotation radius, center, and axis direction,
the real robot performed the door-opening using the learned policy.
Although the policy was trained only using simulated data,
the policy is directly applicable to the real robot as
the sufficient goal state does not require observability on the real robot
and because the policy action decisions only rely on the states with very small sim-to-real gaps.

\subsection{execution}

\figref{fig:re}-(A) shows the task-sequencing simulator used with a demonstrated sequence by a human.
Instead of a configured sequence as in the previous training experiments, the sequence is automatically generated through
demonstration decomposition using the method described in \cite{wake2021verbal}.
The same concept models from the training experiments are used with the policy-update being disabled
(the simulator is not connected to any training algorithm and instead uses a fixed learned policy without updates).

\figref{fig:re}-(B) shows a simulated execution of the demonstrated sequence.
The first row shows the outputs of the physics engine and the second row shows the outputs of the rendering engine.
As both training and execution run on the same system, the learned policy can easily be used as a simulation-for-execution.
The learned policy is already a building block that can be combined with other tasks to generate an application
such as ``pick up a cup from the upper shelf and re-place it to the bottom shelf."

\figref{fig:re}-(C) shows a real robot execution of the demonstrated sequence
by switching the engines in the environment engine pipeline to connect with ROS.
This shows how the simulator can go from the simulated robot execution to the real robot execution
by using the same policy connections but by changing the engines in which the states are obtained, and the actions are performed against.
Usually, going from simulation to real introduces a sim-to-real gap.
However, only part of the scenario sequence uses a learned policy and due to the careful design of the concept models
to divide learning observable dynamics (necessary goal states) from learning hidden dynamics (sufficient goal states),
no such gap was encountered.

\figref{fig:re}-(D) shows an execution of a different sequence ``pick up a cup from the table and throw it in the trash."
This scenario uses the same concept models and only differs in the demonstrated input,
showing how using the simulator and concept model descriptions enable reusing the learned policies for different execution scenarios.
If a policy was learned against a full ``pick-and-place" scenario, the policy would not easily scale to the ``pick-and-throw" scenario
as the problem dynamics are different.
\section{Conclusions}
\label{conclusions}

This article introduced the task-sequencing simulator which bridges simulation-for-learning to simulation-for-execution.
The simulation scenario for learning is created using a sequence of tasks.
This way the simulation-for-learning has the same structure as the simulation-for-execution.
At its core, the simulator uses a concept model which enables sequencing mixed programmed, trained, and under-training building blocks.
While the simulator has a large advantage in terms of integrated system development,
the simulator also provides new directions for simulation in execution and simulation in learning.

From an execution perspective, the simulator allows composing a task-sequence using both programmed and trained tasks.
Unlike programmed-only sequences, the advantage of mixing trained blocks is that, some of the tasks can contain uncertainty and
the goal state of a task can be described using implicit system parameters (the goal state does not have to be obtained directly from the real robot).
The key is that, whether the observed state and selected actions suffice the goal state is learned as a termination signal through training.

From a learning perspective, the simulator and concept model design have the following advantages:
First, the simulation is reusable and easily applicable to slight changes in the scenario.
A policy for a different end-effector can be learned by just changing the actor configurations in the concept model.
A policy can be optimized for different scenarios by just changing the subsequent task in the sufficient goal state.
Second, defining the learning problem using the concept model design enables
a hierarchical learning-structure as well as a structure for reducing sim-to-real gaps.
Any state parameters that do not have a large gap when observed with the real robot is used for defining the necessary goal state,
whereas any state parameters that have a large gap when observed with the real robot is a sufficient goal state
(implicitly learned in simulation but no need to be observed with the real robot).
This type of formulation is possible as only the parts with uncertainty are being learned instead of learning the entire scenario sequence.
Following this structured formulation has allowed going from simulation to real without any extra real-world data collection
and achieving a reusable policy applicable to different execution scenarios.



%




\section*{Acknowledgment}
The authors thank Brice Chung's team, Aydan Aksoylar and Kartavya Neema for their help in the
reward designs and training of the concept models used in the experiments.





%



\ifconfletter
\bibliographystyle{IEEEtran}
\else
\bibliographystyle{unsrt}
\fi
\bibliography{bib}

%








\end{document}